\documentclass[conference]{IEEEtran}
\IEEEoverridecommandlockouts
\usepackage{graphicx}
\usepackage{latexsym}
\usepackage{subfigure}
\usepackage{multirow}
\usepackage{comment}
\usepackage{amsmath}
\usepackage{amsthm}
\usepackage{amssymb}
\usepackage{multirow}
\usepackage{microtype}
\usepackage{booktabs}
\usepackage{cite}

\usepackage{mathtools}

\theoremstyle{plain}
\newtheorem{theorem}{Theorem}
\theoremstyle{definition}
\newtheorem{definition}[theorem]{Definition}

\def\BibTeX{{\rm B\kern-.05em{\sc i\kern-.025em b}\kern-.08em
    T\kern-.1667em\lower.7ex\hbox{E}\kern-.125emX}}

\makeatletter
 
\def\hlinewd#1{%
\noalign{\ifnum0=`}\fi\hrule \@height #1 \futurelet
\reserved@a\@xhline}
\makeatother

\author{Haruki Yokota$^1$ \quad Hiroshi Higashi$^{1}$ \quad Yuichi Tanaka$^{1}$ \quad Gene Cheung$^{2}$ \\
$^1$ \textit{Osaka University, Osaka, Japan} \\
 $^{2}$ \textit{York University, Toronto, Canada}

}
    
\begin{document}

\title{Efficient Learning of Balanced Signed Graphs via Iterative Linear Programming \thanks{The work of H. Yokota is supported by JST SPRING JPMJSP2138.}\thanks{The work of Y. Tanaka is supported in part by JSPS KAKENHI (23H01415, 23K17461) and JST AdCORP JPMJKB2307.}
\thanks{The work of G. Cheung was supported in part by the Natural Sciences and Engineering Research Council of Canada (NSERC) RGPIN-2019-06271, RGPAS-2019-00110.}}

\maketitle

\begin{abstract}
Signed graphs are equipped with both positive and negative edge weights, encoding pairwise correlations as well as anti-correlations in data. 
A balanced signed graph has no cycles of odd number of negative edges. 
Laplacian of a balanced signed graph has eigenvectors that map simply to ones in a similarity-transformed positive graph Laplacian, thus enabling reuse of well-studied spectral filters designed for positive graphs.
We propose a fast method to learn a balanced signed graph Laplacian directly from data. 
Specifically, for each node $i$, to determine its polarity $\beta_i \in \{-1,1\}$ and edge weights $\{w_{i,j}\}_{j=1}^N$, we extend a sparse inverse covariance formulation based on linear programming (LP) called CLIME, by adding linear constraints to enforce ``consistent" signs of edge weights $\{w_{i,j}\}_{j=1}^N$ with the polarities of connected nodes---i.e., positive/negative edges connect nodes of same/opposing polarities. 
For each LP, we adapt projections on convex set (POCS) to determine a suitable CLIME parameter $\rho > 0$ that guarantees LP feasibility.
We solve the resulting LP via an off-the-shelf LP solver in $\mathcal{O}(N^{2.055})$. 
Experiments on synthetic and real-world datasets show that our balanced graph learning method outperforms competing methods and enables the use of spectral filters and graph neural networks (GNNs) designed for positive graphs on balanced signed graphs.
\end{abstract}

\section{Introduction}
\label{sec:intro}

Modern data with graph-structured kernels can be processed using analytical \textit{graph signal processing} (GSP) tools such as graph transforms and wavelets \cite{shuman_emerging_2013,ortega_graph_2018,tanaka_sampling_2020} or deep-learning (DL)-based \textit{graph neural networks} (GNNs) \cite{kipf17}.  
A basic premise in graph-structured data processing is that a finite graph capturing pairwise relationships is available \textit{a priori}; if such graph does not exist, then it must be learned from observable data---a problem called \textit{graph learning} (GL). 

A wide variety of GL methods exist in the literature based on signal smoothness, statistics, and diffusion kernels \cite{kalofolias_how_2016,friedman_sparse_2008a,Thanou_learning_2016,mateos2019,dong2019,bagheri2024}. 
However, most methods focus on learning \textit{unsigned} positive graphs, \textit{i.e.}, graphs with positive edge weights that only encode pairwise correlations between nodes. 
As a consequence, most developed graph spectral filters \cite{onuki_graph_2016,pang17,shuman20} and GNNs \cite{kipf17} are applicable only to positive graphs.
This is understandable, as the notion of graph frequencies is well studied for positive graphs---\textit{e.g.}, eigen-pairs $(\lambda_i,\mathbf{v}_i)$ of the combinatorial graph Laplacian matrix $\mathbf{L}$ are commonly interpreted as graph frequencies and Fourier modes respectively \cite{ortega_graph_2018}, and spectral graph filters with well-defined frequency responses are subsequently designed for signal restoration tasks such as denoising, dequantization, and interpolation \cite{pang17,liu17,chen24}. 

In many practical real-world scenarios, there exist data with inherent pairwise \textit{anti-correlations}.
An illustrative example is voting records in the US Senate, where Democrats / Republicans typically cast opposing votes, and thus edges between them are more appropriately modeled using negative weights. 
The resulting structure is a \textit{signed graph}---a graph with both positive and negative edge weights \cite{hou_laplacian_2003,wu_spectral_2011,kunegis_spectral_2010,dittrich_signal_2020}. 
However, the spectra of signed graph variation operators, such as adjacency and Laplacian matrices, are not well understood.
As a result, designing spectral filters for signed graphs remains difficult.

One exception is \textit{balanced} signed graphs. 
A signed graph is balanced if there exist no cycles of odd number of negative edges \cite{harary_notion_1953}. 
It is discovered that there exists a simple one-to-one mapping from eigenvectors $\mathbf{U}$ of a balanced signed (generalized) graph Laplacian matrix $\mathcal{L}^b$ to eigenvectors $\mathbf{V} = \mathbf{T} \mathbf{U}$ of a similarity-transformed positive graph Laplacian matrix $\mathcal{L}^+ = \mathbf{T} \mathcal{L}^b \mathbf{T}^{-1}$ \cite{yang_signed_2022}, where $\mathbf{T}$ is a diagonal matrix with entries $T_{i,i} \in \{1,-1\}$.
Thus, the spectrum of a balanced signed graph $\mathcal{G}^b$ is equivalent to the spectrum of the corresponding positive graph $\mathcal{G}^+$, and developed spectral filters for positive graphs can be reused for balanced signed graphs.
However, existing GL methods computing balanced signed graphs are limited to sub-optimal two-step approaches\footnote{The exception is a recent work \cite{matz23}, which employs a complex definition of signed graph Laplacian requiring the absolute value operator \cite{dittrich_signal_2020}. We show that our scheme noticeably exceeds \cite{matz23} in performance in Section\;\ref{sec:results}.}: first compute a signed graph from data using, for example, \textit{graphical lasso} (GLASSO) \cite{mazumder_graphical_2012}, then balance the computed graph via ad-hoc and often computation-expensive balancing algorithms
\cite{akiyama_balancing_1981,dinesh_lineartime_2022,yokota_signed_2023}.

\begin{figure}[tb]
\centering
\includegraphics[width = 0.46\textwidth]{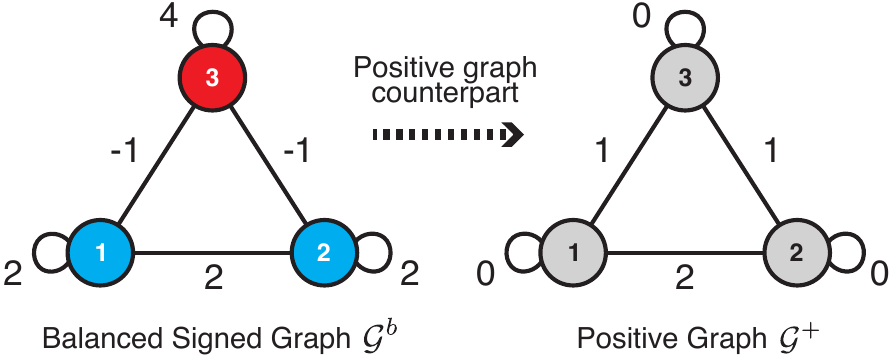}
\vspace{-0.1in}
\caption{Balanced signed graph $\mathcal{G}^b$ (left) and its similarity-transformed positive graph $\mathcal{G}^+$ (right). Numbers inside node denote node indices, and numbers beside edges denote edge weights. Blue-/red-colored nodes have positive/negative polarities, respectively.}
\label{fig:balanced_vs_positive}
\end{figure}

In this paper, we propose a fast GL method that computes a balanced signed graph \textit{directly} from observed data.
Specifically, we extend a previous sparse inverse covariance formulation based on linear programming (LP) called CLIME \cite{Cai_aconstrained_2011} to compute a balanced signed graph Laplacian $\mathcal{L}^b$ given a sample covariance matrix $\mathbf{C}$. 
The \textit{Cartwright-Harary Theorem} (CHT) \cite{harary_notion_1953} states that after appropriately assigning \textit{polarities} $\beta_i \in \{1, -1\}$ to each graph node $i$, a balanced signed graph $\mathcal{G}^b$ has positive / negative edges connecting node-pairs of same / opposing polarities, respectively. 
See Fig.\;\ref{fig:balanced_vs_positive}(left) for an illustration of a balanced signed graph $\mathcal{G}^b$ with nodes assigned with suitable polarities.
\textit{The key idea is to add sign constraints on edge weights $\{w_{i,j}\}_{j=1}^N$ when node $i$ assumes polarity $\beta_i$ during optimization to maintain graph balance.}

After adding edge weight sign constraints, the posed LP (solvable in $\mathcal{O}(N^{2.055})$ using an existing LP solver \cite{jiang_faster_2020}) for node $i$ to determine its polarity $\beta_i$ and edge weights $\{w_{i,j}\}_{j=1}^N$ may not feasible for a chosen CLIME parameter $\rho$ that specifies a target sparsity level in $\mathcal{L}^b$. 
For each LP, we adapt a variant of \textit{projections on convex sets} (POCS) \cite{kacmarz37,escalante2011alternating} to determine a suitable $\rho$ in $\mathcal{O}(N)$ to ensure LP feasibility. 
Experiments show that our method constructs better quality signed graph Laplacians than previous schemes \cite{dinesh_lineartime_2022,yokota_signed_2023,matz23}, while enabling reuse of graph filters \cite{pesenson_variational_2009} and GNNs designed for positive graphs for tasks such as signal denoising.

\vspace{0.02in}
\noindent
\textbf{Notation:}
Vectors and matrices are written in bold lowercase and uppercase letters, respectively.
The $(i,j)$ element and the $j$-th \textit{row} of a matrix $\mathbf{A}$ are denoted by $A_{i,j}$ and $\mathbf{A}_{j}$, respectively.
The $i$-th element in the vector $\mathbf{a}$ is denoted by $a_{i}$.
The square identity matrix of rank $N$ is denoted by $\mathbf{I}_N$, the $M$-by-$N$ zero matrix is denoted by $\mathbf{0}_{M,N}$, and the vector of all ones / zeros of length $N$ is denoted by $\mathbf{1}_N$ / $\mathbf{0}_N$, respectively.
Operator $\|\cdot\|_p$ denotes the $\ell_p$-norm. 

\vspace{-0.02in}
\section{Preliminaries}
\label{sec:prelim}

\vspace{-0.02in}
\subsection{Signed Graphs}
\label{sec:signed_graphs}

\vspace{-0.02in}
Denote by $\mathcal{G} (\mathcal{V},\mathcal{E},\mathbf{W})$ an undirected signed graph with node set $\mathcal{V} = \{1, \dots, N\}$, edge set $\mathcal{E}$, and symmetric weighted adjacency matrix $\mathbf{W} \in \mathbb{R}^{N \times N}$, where weight $w_{i,j} \in \mathbb{R}$ of edge $(i,j) \in \mathcal{E}$ connecting nodes $i, j \in \mathcal{V}$ is $W_{i,j}$. 
We assume that $W_{i,j}, \forall i|i\neq j$, can be positive/negative to encode pairwise correlation / anti-correlation, while self-loop weight $W_{i,i}$ is non-negative, \textit{i.e.}, $W_{i,i} \geq 0, \forall i$. 
Denote by  $\mathbf{D} \in \mathbb{R}^{N \times N}$ a diagonal \textit{degree matrix}, where $D_{i,i} = \sum_{j} W_{i,j}$. 
We define the standard \textit{generalized graph Laplacian matrix}\footnote{We use the standard generalized graph Laplacian for signed graphs instead of the signed graph Laplacian definition in \cite{dittrich_signal_2020} that requires the absolute value operator. This leads to a simple mapping of eigenvectors of a balanced signed graph Laplacian to ones of a similarity-transformed positive graph Laplacian, and a statistics-driven GL problem formulation extending from \cite{Cai_aconstrained_2011}.} for signed graph $\mathcal{G}$ as $\mathcal{L} \triangleq \mathbf{D} - \mathbf{W} + \operatorname{diag}(\mathbf{W})$.

\subsection{Balanced Signed Graphs}
\label{sec:balanced_graphs}

A signed graph is balanced if there are no cycles of odd number of negative edges \cite{harary_notion_1953}. 
We rephrase an equivalent definition of balanced signed graphs, known as the \textit{Cartwright-Harary Theorem} (CHT), as follows.
\begin{theorem}
    \textit{A given signed graph is balanced if and only if its nodes can be polarized into $1$ and $-1$, such that a positive edge always connects nodes of the same polarity, and a negative edge always connects nodes of opposing polarities.}
\end{theorem}
\noindent 
The edges of a balanced signed graph are \textit{consistent}.
\begin{definition}
\label{def:consistent}
\textit{A consistent edge is a positive edge connecting two nodes of the same polarity, or a negative edge connecting two nodes of opposing polarities.}
\end{definition}

For a balanced signed graph $\mathcal{G}^b$ with Laplacian $\mathcal{L}^b = \mathbf{D}^b - \mathbf{W}^b + \text{diag}(\mathbf{W}^b)$, we define its positive graph counterpart $\mathcal{G}^+$ with $\mathcal{L}^+ \triangleq \mathbf{D}^b-|\mathbf{W}^b|+\operatorname{diag}(\mathbf{W}^b)$, where $|\mathbf{W}|$ denotes a matrix with element-wise absolute value of $\mathbf{W}$.
Positive graph $\mathcal{G}^+$ has its own adjacency matrix\footnote{The use of a self-loop of weight equaling to twice the sum of connected negative edge weights is also done in \cite{su17}.} $\mathbf{W}^+$, where $W_{i,j}^+ = |W_{i,j}^b|, \forall i \neq j$, and $W_{i,i}^+ = W_{i,i}^b - 2 \sum_{j} [-W_{i,j}^b]_+$, where $[c]_+$ is a positivity function that returns $c$ if $c \geq 0$ and $0$ otherwise.
Diagonal degree matrix $\mathbf{D}^+$ for $\mathcal{G}^+$ is defined conventionally, where $D_{i,i}^+ = \sum_j W_{i,j}^+$.
Note that adjacency matrix $\mathbf{W}^+$ is defined so that the generalized Laplacian is given as
\begin{align}
\mathcal{L}^+ = \mathbf{D}^+ - \mathbf{W}^+ + \text{diag}(\mathbf{W}^+) .
\end{align}

In Fig.\;\ref{fig:balanced_vs_positive}, we illustrate an example of a balanced signed graph $\mathcal{G}^b$ and its positive graph counterpart $\mathcal{G}^+$. 
In the example, the weights of edges incident on node $3$ are $W^b_{1,3} = W^b_{2,3} = -1$, and the self-loop has weight $W^b_{3,3} = 4$. 
For the positive graph counterpart, the self-loop on node $3$ is $W_{3,3}^+ = 4 - 2(1+1) = 0$.

Laplacian $\mathcal{L}^b = \mathbf{U} \boldsymbol{\Lambda} \mathbf{U}^\top$ of $\mathcal{G}^b$ enjoys a one-to-one mapping of its eigenvectors to those of its positive graph counterpart $\mathcal{L}^+$ \cite{yang_signed_2022}.
Specifically, denote by $\mathcal{V}_1$ and $\mathcal{V}_{-1}$ the node subsets in $\mathcal{G}^b$ with polarity $1$ and $-1$ respectively.
We reorder rows / columns of balanced signed graph Laplacian $\mathcal{L}^b$ so that nodes in $\mathcal{V}_{1}$ of size $M$ are indexed before nodes in $\mathcal{V}_{-1}$ of size $N-M$. 
Then, using the following invertible diagonal matrix 
\begin{align}
\mathbf{T} &= \left[ \begin{array}{cc}
\mathbf{I}_M & \mathbf{0}_{M,N-M} \\
\mathbf{0}_{N-M,M} & -\mathbf{I}_{N-M}
\end{array} \right], 
\label{eq:matrixT}
\end{align}
we see that $\mathcal{L}^b$ and $\mathcal{L}^+$ are \textit{similarity transform} of each other: 
\begin{align}
\mathbf{T} \mathcal{L}^b \mathbf{T}^{-1} &\stackrel{(a)}{=} \mathbf{T} \mathbf{U} \boldsymbol{\Lambda} \mathbf{U}^\top \mathbf{T}^\top \stackrel{(b)}{=} \mathbf{V} \boldsymbol{\Lambda} \mathbf{V}^\top 
\\
&\stackrel{(c)}{=} \mathbf{T} \left( \mathbf{D} - \mathbf{W} + \text{diag}(\mathbf{W}) \right) \mathbf{T} \\
&= \mathbf{D} - |\mathbf{W}| + \text{diag}(\mathbf{W}) = \mathcal{L}^+
\end{align}
where in $(a)$ we write $\mathbf{T}^\top = \mathbf{T}^{-1} = \mathbf{T}$, in $(b)$ $\mathbf{V} \triangleq \mathbf{T} \mathbf{U}$, and in $(c)$ we apply the definition of generalized graph Laplacian.
Thus, $\mathcal{L}^+$ and $\mathcal{L}^b$ have the same set of eigenvalues, and their eigen-matrices are related via matrix $\mathbf{T}$. 
For the example in Fig.\;\ref{fig:balanced_vs_positive}, $\mathcal{L}^+$ relates to $\mathcal{L}^b$ via $\mathbf{T} = \text{diag}([1,1,-1])$.

\vspace{0.05in}
\noindent
\textit{Spectral Graph Filters}: 
Spectral graph filters are frequency modulating filters, where graph frequencies are conventionally defined as the \textit{non-negative} eigenvalues of a positive graph Laplacian matrix \cite{cheung18}. 
Thus, to reuse spectral filters designed for positive graphs \cite{shuman20} for a balanced signed graph $\mathcal{G}^b$, it is sufficient to require $\mathcal{L}^b$ to be \textit{positive semi-definite} (PSD), since similarity-transformed $\mathcal{L}^+$ shares the same eigenvalues. 

\subsection{Sparse Inverse Covariance Learning}
\label{sec:clime}

Given a data matrix $\mathbf{X}=[\mathbf{x}_1,\dots,\mathbf{x}_K]$ where $K > N$ and $\mathbf{x}_k \in \mathbb{R}^{N}$ is the $k$-th observation, a sparse inverse covariance matrix $\mathcal{L} \in \mathbb{R}^{N\times N}$ (interpretable as a signed graph Laplacian) can be estimated from data via a constrained $\ell_1$-minimization formulation called CLIME \cite{Cai_aconstrained_2011}. 
In a nutshell, CLIME seeks a \textit{positive definite} (PD) $\mathcal{L}$ given a PD sample covariance matrix $\mathbf{C} = \frac{1}{N-1}\mathbf{XX}^\top$ (assuming zero mean) via a \textit{linear programming} (LP) formulation:
\begin{equation}
\label{eq:clime}
\min_\mathcal{L}\quad \|\mathcal{L}\|_1, \quad \text{s.t.}\quad \|\mathbf{C} \mathcal{L}-\mathbf{I}_N\|_\infty\leq \rho
\end{equation}
where $\rho \in\mathbb{R}_+$ is a parameter specifying a target sparsity level. 
Specifically, \eqref{eq:clime} computes a sparse $\mathcal{L}$ (promoted by the $\ell_1$-norm objective) that approximates the right inverse of $\mathbf{C}$. 
The $i$-th column $\mathbf{l}_i$ of $\mathcal{L}$ can be computed independently:
\begin{equation}
\label{eq:clime_i}
\min_{\mathbf{l}_i}\quad \|\mathbf{l}_i\|_1,\quad\text{s.t.}\quad\|\mathbf{C} \, \mathbf{l}_i-\mathbf{e}_i\|_{\infty}\leq\rho 
\end{equation}
where $\mathbf{e}_i$ is the canonical vector with one at the $i$-th entry and zero elsewhere. 
\eqref{eq:clime_i} can be solved using an off-the-shelf LP solver such as \cite{jiang_faster_2020} with complexity $\mathcal{O}(N^{2.055})$. 
The resulting $\mathcal{L} = [\mathbf{l}_1,\mathbf{l}_2,\dots,\mathbf{l}_N]$ is not symmetric in general, and \cite{Cai_aconstrained_2011} computes a symmetric approximation $\mathcal{L}^* \triangleq (\mathcal{L} + \mathcal{L}^\top) / 2$ as a post-processing step. 

\section{Balanced Signed Graph Formulation}
\label{sec:balanced}

\subsection{Optimization Approach}
\label{subsec:approach}

Our goal is to estimate a sparse, PD, balanced signed graph Laplacian $\mathcal{L}^b\in\mathbb{R}^{N\times N}$ directly from empirical covariance matrix $\mathbf{C}$. 
Our algorithm focuses on one node $i$ at a time, optimizing its polarity $\beta_i \in \{-1,1\}$ and weights $\{w_{i,j}\}_{j=1}^N$ of edges to other nodes $j$. 
CHT states that in a balanced signed graph, same-/opposing-polarity node pairs are connected by edges of positive/negative weights.
Given the edge sign restrictions, for each node $i$, we execute a variant of optimization \eqref{eq:clime_i} \textit{twice}, where the signs of the entries in variable $\mathbf{l}_i$ (edge weights $\{w_{i,j}\}_{j=1}^N$) are restricted for consistency, given an assumed polarity $\beta_i$.  
We then select node $i$'s polarity $\beta_i$ corresponding to the smaller of the two objective values. 

Specifically, our algorithm is
\begin{enumerate}
\item Initialize polarities $\beta_j$ for all graph nodes $j$.
\item For each column $\mathbf{l}_i$ of $\mathcal{L}^b$ corresponding to node $i$,
\begin{enumerate}
    \item Assume node $i$'s polarity $\beta_i =1 $ or $\beta_i = -1$.
    \item Optimize $\mathbf{l}_i$ given $\mathbf{C}$, ensuring $\text{sign}(l_{i,j})$ are consistent with the polarities of connected nodes $j$.
    \item Select polarity $\beta_i$ for node $i$ with the smaller objective value. 
    \item Update $i$-th column / row of $\mathcal{L}^b$ using $\mathbf{l}_i$ of node $i$ with polarity $\beta_i$.
\end{enumerate}
\item Update columns / rows of $\mathcal{L}^b$ until convergence.
\end{enumerate}
Note that by simultaneously updating the $i$-th column / row of $\mathcal{L}^b$ in step 2(d), $\mathcal{L}^b$ remains symmetric.

\subsection{Linear Constraints for Consistent Edges}

To augment CLIME for a balanced signed graph Laplacian, we first construct linear constraints that ensure consistent edges. 
From Definition\;\ref{def:consistent}, we write the following relationship for an edge $(i,j)\in \mathcal{E}$ of the targeted balanced graph $\mathcal{G}^b$ w.r.t. node polarities $\beta_i$ and $\beta_j$:
\begin{equation}
\label{eq:consistent}
    \beta_i\beta_j\operatorname{sign}(W_{i,j})=
\begin{cases}1 & \text{if }(i,j)\text{ is consistent,}\\-1 & \text{if }(i,j)\text{ is inconsistent.}\end{cases}
\end{equation}
For Laplacian entries $\mathcal{L}_{i,j}^b$, linear constraints for edge consistency can be written as
\begin{equation}
\label{eq:b_const}
    \beta_i\beta_j \mathcal{L}_{i,j}^b \leq 0, ~~~\forall j \,|\, j \neq i .
\end{equation}

\subsection{Optimization of Laplacian Column}
\label{sec:prop}

We reformulate LP \eqref{eq:clime_i} by incorporating linear constraints \eqref{eq:b_const}.
Denote by $\mathbf{S}^i \in\mathbb{R}^{N \times N}$ a diagonal matrix, where diagonal entries $S_{j,j}^i = \beta_i \beta_j, \forall j|j \neq i$ and $S_{i,i}^i = -1$. 
Optimization \eqref{eq:clime_i} becomes
\begin{equation}
\label{eq:prop2}
\min_{\mathbf{l}_i} \|\mathbf{l}_i\|_1 \quad\text{s.t.} \quad \begin{cases} \|\mathbf{C} \mathbf{l}_i-\mathbf{e}_i \|_\infty\leq\rho\\\mathbf{S}^i \mathbf{l}_{i}\leq \mathbf{0}_N \end{cases} .
\end{equation}
The vector inequality means entry-wise inequality. 
By incorporating one additional linear constraint, \eqref{eq:prop2} remains an LP.

Unlike \eqref{eq:clime_i} that always has at least one feasible solution for any $\rho > 0$ (\textit{i.e.}, $i$-th column of $\mathbf{C}^{-1}$), the additional inequality constraint in \eqref{eq:prop2} means that the LP may be \textit{infeasible}. 
Thus, the selection of $\rho$ to enable a feasible solution for at least one of two possible polarities $\beta_i$ of node $i$ becomes crucial.

\subsection{Selection of $\rho$ via POCS}
\label{sec:opt}
\begin{table*}[tb]
	\centering
	\caption{Synthetic Experiment Results ($N=50, K=500$)}\vspace{-5pt}
	\label{tab:synth}
	\begin{tabular}{ c | c | c c c | c c c } \hline
         & Proposed & \multicolumn{3}{c|}{GLASSO} & \multicolumn{3}{c}{CLIME} \\ \hline
        &-&Min&Max&Greed&Min&Max&Greed\\ \hline
        FM $\uparrow$ & \bf{0.6679}&0.4841&0.4821&0.6117&0.4983& 0.4938& 0.6405\\\hline
        RE $\downarrow$ &\bf{0.2854}&0.4055 &0.4055&0.3705& 0.3335&0.3343&0.2898 \\ \hline
        \end{tabular}\vspace{-10pt}
 \label{table:1}
\end{table*}

\begin{table*}[h]
\caption{Denoising Results (Air Pressures in Japan, $N = 96, K = 2016$)}\vspace{-5pt}
\label{tab:real}
	\centering
	\begin{tabular}{ c | c c | c c || c c | c c} \hline
        Methods &\multicolumn{2}{|c|}{Low Pass Filter}&\multicolumn{2}{|c||}{Wavelet Filter}&\multicolumn{2}{|c|}{GCN}&\multicolumn{2}{|c}{GAT}\\\hline
        Noise Level & $\sigma=0.20$ & $\sigma=0.25$ &$\sigma=0.20$ & $\sigma=0.25$ &$\sigma=0.20$ & $\sigma=0.25$ & $\sigma=0.20$ & $\sigma=0.25$\\ \hline
	Proposed & \bf{0.2540} & \bf{0.2645} & \bf{0.1670} &  \bf{0.1929} & \bf{0.0617} & \bf{0.0691} & \bf{0.0665} & \bf{0.0727}\\ \hline
        CLIME-Greed    & 0.2703 & 0.2812 & {0.1823}  & 0.2136 & 0.0747 & 0.0810 & 0.0691 & 0.0737 \\ \hline
        GLASSO-Greed   & 0.4175 & 0.4160 & {0.2177} & 0.2513 & 0.0659 & 0.0746 & 0.0677 & 0.0753\\ \hline
        BSigGL   & 0.3455 & 0.4129 & {0.2059} & 0.2577 & 0.0761 & 0.0841 & 0.0857 & 0.0944\\ \hline
        GGL      & 0.3629 & 0.3647 & {0.2061} & 0.2349 & {0.0620} & {0.0711} & 0.0663 & 0.0745 \\\hline 
	\end{tabular}\vspace{-15pt}
\end{table*}
For each node $i$, we seek to select $\rho > 0$ so that LP \eqref{eq:prop2} has at least one feasible solution $\mathbf{l}_i$ for at least one polarity $\beta_i \in \{-1, 1\}$. 
We solve the LP feasibility problem efficiently using one variant of \textit{projections on convex sets} (POCS) \cite{kacmarz37}: given convex sets $\mathcal{S}_1, \ldots \mathcal{S}_S$ that are \textit{half-spaces}\footnote{A half-space $\mathcal{S}$ is one ``half" of the partitioned $N$-dimensional Euclidean space, defined by $\mathcal{S} = \{\mathbf{x} \in \mathbb{R}^N \,|\, \mathbf{c}^\top \mathbf{x} \leq c_0\}$, where $\mathbf{c}^\top \mathbf{x} = c_0$ is a hyperplane parameterized by $\mathbf{c} \in \mathbb{R}^N$ and $c_0 \in \mathbb{R}$.}, repeated cyclical applications of respective linear projections $\text{proj}_{\mathcal{S}_1}(\cdot), \ldots, \text{proj}_{\mathcal{S}_S}(\cdot)$ will converge to an intersection point $\mathbf{p} \in \mathcal{S}_1 \cap \cdots \cap \mathcal{S}_S$ if one exists. 
If no intersection point exists, then the same closest points in sets $\mathcal{S}_1, \ldots, \mathcal{S}_S$ will repeat.

To apply POCS, given the two constraints in \eqref{eq:prop2}, we first rewrite them as $3N$ linear constraints:
\begin{align}
\mathbf{C}_j \mathbf{l}_i &\leq \rho, ~~ \forall j|j \neq i,  & \mathbf{C}_i \mathbf{l}_i \leq 1 + \rho   \nonumber\\
\mathbf{C}_j \mathbf{l}_i &\geq \rho, ~~ \forall j|j \neq i, & \mathbf{C}_i \mathbf{l}_i \geq 1 - \rho \nonumber\\
\mathbf{S}^i_j \mathbf{l}_i &\leq \mathbf{0}_N, ~~\forall j
\label{eq:halfspace}
\end{align}
where $\mathbf{C}_j$ and $\mathbf{S}^i_j$ denote the $j$-th rows of matrix $\mathbf{C}$ and $\mathbf{S}^i$, respectively. 
Each constraint in \eqref{eq:halfspace} defines a half-space $\mathcal{S}$ in the form $\mathbf{c}^\top \mathbf{x} \leq c_0$. 
To project a candidate solution $\tilde{\mathbf{l}}_i$ into $\mathcal{S}$, we first check if $\mathbf{c}^\top \tilde{\mathbf{l}}_i \leq c_0$.
If so, $\tilde{\mathbf{l}}_i \in \mathcal{S}$ already. 
If not, we project $\tilde{\mathbf{l}}_i$ onto the hyperplane $\mathbf{c}^\top \mathbf{x} = c_0$ that defines $\mathcal{S}$:
\begin{align}
\mathbf{l}^*_i &= \left(\mathbf{I}_N - \frac{\mathbf{c} \mathbf{c}^\top}{\mathbf{c}^\top \mathbf{c}} \right) \tilde{\mathbf{l}}_i + \frac{c_0}{\mathbf{c}^\top \mathbf{c}} \mathbf{c} .
\end{align}

Starting from an initialized $\rho$, we increase $\rho$ slowly until POCS confirms LP feasibility in $\mathcal{O}(N)$. 
An existing LP solver \cite{jiang_faster_2020} then solves the feasible LP in $\mathcal{O}(N^{2.055})$.

\vspace{-0.02in}
\section{Experiments}
\label{sec:results}

\vspace{-0.02in}
We present the results of balanced signed graph learning on both synthetic data and denoising of real data.

\textit{Synthetic Data:} We randomized a graph based on Erd\"{o}s--R\'{e}nyi (ER) model \cite{Erdos1959pmd} with $N=50$ nodes and edge probability of $p=0.2$. 
Edge weight magnitudes were set randomly in the uniform range $[0.01, 1]$. 
We randomized each node $i$'s polarity $\beta_i \in \{1, -1\}$ with equal probability. 
Edge weight signs were set to positive/negative for each node-pair with same/opposing polarities, resulting in a balanced graph.
Self-loop on each node was set as $w_{i,i} = 2.5\sum_j [-w_{i,j}]_+$, to ensure the resulting graph Laplacian $\mathcal{L}^b$ is PD and invertible.
$K=500$ signal observations were generated following the Gaussian Markov Random Field (GMRF) model as $\mathbf{x}^{k} \sim \mathcal{N}(\mathbf{0}_{N},(\mathcal{L}^b)^{-1})$.

\textit{Hourly Air Pressures in Japan\footnote{https://www.data.jma.go.jp/stats/etrn/index.php} (APJ):} This dataset consisted of hourly air pressure records from $96$ weather stations in Japan from March 2022 to May 2022. 
The total observation number was $K = 2016$, and we computed a moving average for every $6$ hours. 
The observations at each node were normalized, then contaminated with \textit{additive white Gaussian noise} (AWGN) with noise levels $\sigma=\{0.20, 0.25\}$. 
\noindent\textbf{Synthetic Data Experiment:}
We compare the performance of our algorithm against variants of conventional two-step balancing approaches: a precision matrix estimation step followed by a graph balancing step. 
We employed CLIME \cite{Cai_aconstrained_2011} or Graphical Lasso (GLASSO) \cite{mazumder_graphical_2012} for precision matrix estimation. 
The three variants of graph balancing methods were:
a) \textit{MinCut Balancing (Min)} \cite{yokota_signed_2023}: Nodes were polarized based on the min-cut of positive edges, and inconsistent edges were removed. 
b) \textit{MaxCut Balancing (Max)} \cite{yokota_signed_2023}: Nodes were poloarized based on the max-cut of negative edges, and inconsistent edges were removed. 
c) \textit{Greedy Balancing (Greed)} \cite{dinesh_lineartime_2022}: A polarized set was initialized with a random node polarized to $1$, then nodes connected to the set were greedily polarized one-by-one, so that the number of consistent edges to the set was maximized in each polarization. %

\noindent\textbf{Result:}
We numerically evaluate the performance using F-measure (FM), and relative error (RE) in Table\;\ref{tab:synth}. 
The values were averaged over 30 runs. 
The results show that our method outperforms conventional two-step methods in both metrics.

\noindent\textbf{Real Data Experiment:} We tested competing methods in signal denoising on the APJ dataset.
For positive graph signal denoising, we employed a bandlimited graph low-pass filter (BL), spectral graph wavelets (SGW) \cite{hammond2011wavelets}, graph convolutional net (GCN), and graph attention net (GAT) \cite{rey2021untrained}. 
We first learned a balanced signed graph Laplacian $\mathcal{L}^b$, which was similarity-transformed (via matrix $\mathbf{T}$ in \eqref{eq:matrixT}) to a positive graph Laplacian $\mathcal{L}^+$ as kernel for the denoising algorithms. 
Each observed noisy signal was also similarity-transformed via $\mathbf{T}$ for processing on positive graph $\mathcal{G}^+$.
For BL, we used a filter that preserves the low-frequency band $[0,0.3 \lambda_\text{max}]$, where $\lambda_\text{max}$ was the maximum eigenvalue of the graph Laplacian. 
For SGW, we designed the Mexican hat wavelet filter bank for range $[0, \lambda_\text{max}]$. 
The number of frames to cover the interval was set to $7$. 
For GCN and GAT, we followed the GNN architecture in \cite{rey2021untrained} for signal denoising on positive graphs.
We also compared our methods against the balanced signature graph learning method (BSigGL) \cite{matz23} and a positive graph learning algorithm (GGL) in \cite{egilmez_graph_2017}. 

\noindent\textbf{Result:}
We summarize the results in Table \ref{tab:real}. 
The values are the average root mean squared errors (RMSE) of the restored signals for filtering methods and normalized MSE for GNN-based methods. 
Results show that the balanced signed graph learned using our method yields the best performance across all denoising schemes designed for positive graphs.

\vspace{-0.02in}
\section{Conclusion}
\label{sec:conclude}

\vspace{-0.02in}
We proposed an efficient algorithm to learn a balanced signed graph directly from data.
We augment a previous sparse inverse covariance matrix formulation based on linear programming (LP) \cite{Cai_aconstrained_2011} with additional linear constraints for graph balance.
We ensure LP feasibility with a suitable selection of a sparsity parameter $\rho$ via a variant of projections on convex sets (POCS) \cite{kacmarz37}.
In experiments, we showed that our balanced graph learning method enables reuse of spectral filters / GNNs for positive graphs and outperforms previous learning methods.

\bibliographystyle{IEEEtran}
\bibliography{IEEEabrv,ref}

\end{document}